\documentclass{article}
\usepackage{spconf,amsmath,graphicx}
\usepackage{amssymb}
\usepackage{booktabs}
\usepackage{multirow}
\usepackage{color}
\usepackage[numbers,sort&compress]{natbib}
\title{Cross-Domain Few-Shot Classification via Inter-Source Stylization}
\name{Huali Xu$^{1}$ \qquad Shuaifeng Zhi$^{2}$ \qquad Li Liu$^{* 2}$ \thanks{*Corresponding author}}
  
  \address{$^{1}$ Center for Machine Vision and Signal Analysis (CMVS), University of Oulu, Oulu, Finland \\
      $^{2}$ College of Electronic Science, National University of Defense Technology, Changsha, China}
      
\begin{document}
%
\maketitle
\begin{abstract}
The goal of Cross-Domain Few-Shot Classification (CDFSC) is to accurately classify a target dataset with limited labelled data by exploiting the knowledge of a richly labelled auxiliary dataset, despite the differences between the domains of the two datasets. Some existing approaches require labelled samples from multiple domains for model training. However, these methods fail when the sample labels are scarce. To overcome this challenge, this paper proposes a solution that makes use of multiple source domains without the need for additional labeling costs. Specifically, one of the source domains is completely tagged, while the others are untagged. An Inter-Source Stylization Network (ISSNet) is then introduced to enhance stylisation across multiple source domains, enriching data distribution and model's generalization capabilities. Experiments on 8 target datasets show that ISSNet  leverages unlabelled data from multiple source data and significantly reduces the negative impact of domain gaps on classification performance compared to several baseline methods.
\end{abstract}
\section{Introduction}
\label{sec:intro}

Few-Shot Classification (FSC) is a problem solving a target task with limited labelled data, exploiting prior knowledge learned from auxiliary datasets with numerous categories and labelled samples. However, most current FSC research assumes that the auxiliary and target datasets are from the same domain, making it challenging to handle the domain shift between the two datasets~\cite{fsl:survey, fsl:classification}.

This has given rise to a more challenging problem called Cross Domain Few-Shot Classification (CDFSC)~\cite{cdfsl,resnet10}, where the auxiliary and target datasets come from two vastly different domains. Existing methods usually address the CDFSC problem from several aspects such as adversarial training~\cite{fsl:adversarial}, model ensembling~\cite{fsl:ensemble}, finetuning~\cite{fsl:mixup}, etc. In these methods, an adversarial learning-to-learning mechanism is introduced in~\cite{cdfsl:Tian} to train the shift layer and generate pseudo tasks. Furthermore,~\cite{fsl:mixup} integrates the proposed mixup module into the meta-learning mechanism. Recently, some attempts start to address CDFSC by introducing supervised samples from multiple different domains into the training phase~\cite{mcd1,mcd2}.~\cite{mcd1} distills knowledge from multiple individually trained networks to learn a unified set of universal deep representations by employing adapters and centered kernel alignment to co-align the features of each network. However, obtaining annotations for these source domains can be costly for real-world applications.

In tackling this issue, this paper proposes a novel solution that leverages multiple source domains, where only one of them is fully labelled ($\mathcal{D}{sl}$) while the rest are left unlabelled ($\mathcal{D}{su}$). In this context, the primary challenge lies in how to choose and effectively use $\mathcal{D}_{su}$ to improve the model generalization. 

\begin{figure}
\begin{center}
\includegraphics[width=0.9\linewidth]{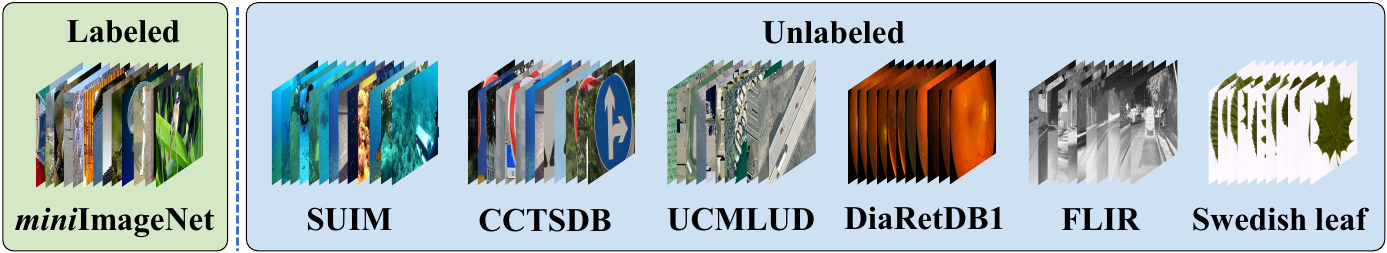}
\end{center}
\vspace{-0.5cm}
   \caption{The multiple source domains include one fully labelled source and six unlabelled sources.}
\label{fig:motivation}
\vspace{-0.5cm} 
\end{figure}

To solve the challenge, this paper first explores a benchmark (as shown in Figure~\ref{fig:motivation}), where \textit{mini}ImageNet is selected as $\mathcal{D}_{sl}$, and the choose of $\mathcal{D}_{su}$ is based on the following principles: (a) they represent a wide range of various practical scenarios and fields, (b) their labelled data are scarce while the unlabelled data are adequate.
In addition, a novel approach, Inter-Source Stylization Network (ISSNet), is proposed to utilize $\mathcal{D}_{su}$ significantly and improves the model's generalization ability by diversifying the domain through style transfer between source domains. 
ISSNet has three components: Inter-Source style transfer learning, Cross-domain distilled consistency learning, and Target prototypical classifier learning. In the Inter-source style transfer learning phase, a style transfer teacher network transfers style information from the unlabelled domains to the labelled domain, expanding its distribution. In the Cross-domain distilled consistency learning phase, the student encoding model and classifier are trained with both the labelled and expanded domains. Finally, in the Target prototypical classifier learning phase, the pre-trained student encoding model is fine-tuned on the target datasets and a new prototypical classifier is learned. The goal of ISSNet is to enhance the model's generalization through multiple source domains, thereby improving the performance of FSC on the target domain. The key contributions of this paper are as follows:

(1) This paper introduces multiple source domains ($\mathcal{D}_{sl}$ and $\mathcal{D}_{su}$) in model training to enhance CDFSC performance without more label costs, where $\mathcal{D}_{sl}$ is fully labelled and $\mathcal{D}_{su}$ is unlabelled.

(2) A new approach ISSNet is presented to improve the CDFSC performance by making full use of the effective information of $\mathcal{D}_{sl}$ and $\mathcal{D}_{su}$. It broadens the distribution of the sources and strengthens the model's generalization.

(3) Experiments on 8 target datasets show that ISSNet effectively reduces the performance loss caused by domain shifts. The 8 datasets include which from near-domain and distance-domain.

\section{Methodology}
\label{sec:3}

\begin{figure}
\begin{center}
\includegraphics[width=0.4\textwidth]{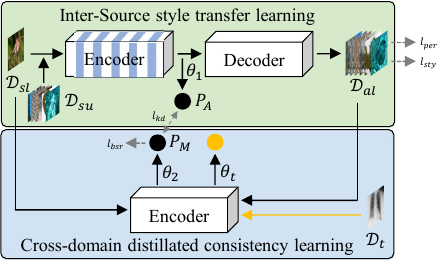}
\end{center}
\vspace{-0.5cm}
   \caption{Overview of ISSNet. Green part shows the Inter-source style transfer learning stage, the blue area indicates the cross-domain distillated consistency learning phase. The target prototypical classifier learning process is represented with yellow line. $\theta$ means the classifier, in which $\theta_{1}$ and $\theta_{2}$ are for sources, and $\theta_{t}$ means the target classifier.}
   \vspace{-0.4cm}
\label{fig:overview}
\end{figure}

This paper presents a new framework, Inter-Source Stylization Network (ISSNet), to transfer the styles of multiple source domains $\mathcal{D}_{su}$ to the labelled source domain $\mathcal{D}_{sl}$ through a style transfer process, thereby expanding the distribution of $\mathcal{D}_{sl}$ and improving the model's generalization, as shown in Figure~\ref{fig:overview}. ISSNet has three stages: (1) Inter-Source style transfer learning, (2) Cross-domain distilled consistency learning, and (3) Target prototypical classifier learning. In the first stage, a teacher style transfer network \textit{A} is trained to transfer the styles from $\mathcal{D}_{su}$ to $\mathcal{D}_{sl}$, generating a new dataset $\mathcal{D}_{al}$, with styles similar to $\mathcal{D}_{su}$ and content from $\mathcal{D}_{sl}$. The encoder of \textit{A} is followed by a classifier to maintain model recognition. In the second stage, $\mathcal{D}_{al}$ and $\mathcal{D}_{sl}$ are input into a student encoding model \textit{M}, followed by a classifier. Knowledge distillation is employed, using a KL loss, to obtain knowledge of \textit{A}. In the last stage, \textit{M} is used to fine-tune and predict on $\mathcal{D}_t$.

\subsection{Inter-Source style transfer learning}
Inter-Source style transfer learning is achieved through network \textit{A} with crossnorm operations~\cite{cnsn}. The goal is to combine the information from $\mathcal{D}_{sl}$ and $\mathcal{D}_{su}$ to generate new pseudo-labelled samples $\mathcal{D}_{al}$. \textit{A} comprises an encoder \textit{$E_{A}$} and a decoder \textit{$D_{A}$} (the structures of \textit{$E_{A}$} and \textit{$D_{A}$} are borrowed from~\cite{adain}). Crossnorm operations are incorporated into every layer of \textit{$E_{A}$}. \textit{$E_{A}$} blends $\mathcal{D}_{sl}$ and $\mathcal{D}_{su}$ and maps to the combined feature $f_{al}$. The equation for the crossnorm operation is as follows:
\begin{equation}
  Crossnorm(f_{sl}, f_{su}) = \sigma(f_{su}) \frac{f_{sl}-\mu(f_{sl})}{\sigma(f_{sl})} + \mu(f_{su}),
  \label{eq:adain}
\end{equation}
where $\mu(\cdot)$ and $\sigma(\cdot)$ indicate the mean value and standard deviation, respectively. Then $f_{al}$ is reconstructed as $\mathcal{D}_{al}$ by \textit{$D_{A}$} while is recognized by a classifier \textit{$C_{A}$}.

The goal of \textit{A} is to optimize the perceptual loss $l_{per}$, style loss $l_{sty}$, and cross entropy (CE) loss $l_{ce}$. $l_{per}$ evaluates the perceptual similarity of two samples by calculating the distance between activation maps of a pre-trained network. It can be expressed as follows:
\begin{equation}
  l_{per} = \mathbb{E} \left[ \sum_{i}\frac{1}{N_{i}} \left \| \phi_{i}\left( \mathcal{D}_{sl} \right) - \phi_{i}\left( \mathcal{D}_{al} \right) \right \|_{1} \right],
  \label{eq:9}
\end{equation}
where $\phi_{i}$ represents the activation map of the $i$th layer of the pre-trained network. We use the VGG-19 network pre-trained on ImageNet~\cite{imagenet} as the pre-trained model, thus $\phi_{i}$ corresponds to the activation maps from layer of the VGG-19 network. Additionally, the style loss $l_{sty}$ measures the discrepancy between covariances of the activation maps, reflecting the difference in textures. $l_{sty}$ can be defined as follows:
\begin{equation}
  l_{sty} = \mathbb{E}_{j}\left[\left \| G_{j}^{\phi}\left( \mathcal{D}_{su} \right) - G_{j}^{\phi}\left( \mathcal{D}_{al} \right) \right \|_{1} \right],
  \label{eq:10}
\end{equation}
where $G_{j}^{\phi}$ is a gram matrix with $C_{j} \times C_{j}$ derived from the activation maps $\phi_{j}$. Additionally, the CE loss $l_{ce}$ is introduced to enhance recognition. The overall objective function $l_{iss}$ is a weighted combination of $l_{ce}$, $l_{per}$, and $l_{sty}$, with weight parameters $\lambda_{per}$ and $\lambda_{sty}$ set to 1 and 10, respectively.
\begin{equation}
  l_{iss} =  l_{ce} + \lambda_{per}l_{per} + \lambda_{sty}l_{sty}.
  \label{eq:11}
\end{equation}

\subsection{Cross-domain distillated consistency learning}
The objective of the cross-domain distillated consistency learning is to encode both $\mathcal{D}_{sl}$ and the pseudo-labelled samples $\mathcal{D}_{al}$ into features $f_{sl}$ and $f_{al}$ using the student encoding model \textit{M}. These features are then fed into a classifier to obtain the corresponding predictions. The prediction of $f_{sl}$ is indicated to \textit{$P_{M}$}. Besides, $f_{sl}$ also is input into \textit{$E_{A}$} and \textit{$C_{A}$} to obtain the corresponding prediction \textit{$P_{A}$}. The knowledge distillation is performed by minimizing the KL divergence between \textit{$P_{M}$} and \textit{$P_{A}$} through the KL loss $l_{kd}$. Additionally, the overall objective loss function includes two CE loss with batch spectral regularization (BSR) $l_{bsr}$ (includes $l_{sl}$ and $l_{al}$ for $\mathcal{D}_{sl}$ and $\mathcal{D}_{al}$, respectively), which regularizes the singular values of the feature matrix in a batch. The objective function can be referred as:
\begin{equation}
  l_{cdd} = l_{kd} + \lambda_{bsr} l_{bsr},
  \label{eq:12}
\end{equation}
where both $\lambda_{bsr}$ is set as 0.05, and $l_{bsr}=l_{sl}+l_{al}$. The $l_{kd}$ is expressed as:
\begin{equation}
  l_{kd} = KL\left( P_{M}, P_{A} \right),
  \label{eq:13}
\end{equation}
where \textit{KL} is the Kullback-Leibler divergence loss, and $l_{bsr}$ is shown as:
\begin{equation}
  l_{bsr}(\textbf{W}) = l_{ce}(\textbf{W}) + \lambda \sum_{i=1}^{n}g_{i}^{2},
  \label{eq:14}
\end{equation}
where $\lambda$ is 0.001, $\textbf{W}$ is the parameters of classifier, and $g_{i}$ $(i=1, 2,..., n)$ are singular values of the batch feature matrix.

\subsection{Target prototypical classifier learning}
The target prototypical classifier learning involves fine-tuning the pre-trained model \textit{M} and training a new classifier for the target dataset. This is achieved by following a few-shot training routine, where \textit{M} is fine-tuned on the target dataset $\mathcal{D}_{t}$ using the \textit{N-way K-shot} setting. Through the fine-tuning process, \textit{M} adapts to the target domain, and a new classifier is trained to categorize the target domain samples. The objective function used to optimize both \textit{M} and the new classifier is the CE loss.

\section{Experiments}
\label{sec:4}
In this section, we first introduce the datasets and experimental setup. Then we demonstrate the experimental results, including the effectiveness of the style transfer network, CDFSC ability analysis. Finally, we performe the ablation study includes the performance of losses and strategies.

\subsection{Datasets and Experimental Setup}
\textbf{Datasets.} This paper improves the model generalization using 7 source datasets from different fields, and evaluate the CDFSC performance on 8 target datasets. The 7 sources include a fully labelled source (\textit{mini}ImageNet~\cite{matchingnet}) and 6 unlabelled sources (DIARETDB1~\cite{eyeball}, Swedish Leaf~\cite{leaf}, FLIR~\footnote{https://www.flir.com/oem/adas/adas-dataset-form/}, SUIM~\cite{SUIM}, CCTSDB~\cite{traffic}, the UC Merced Land Use Dataset (UCMLUD)~\cite{fsl:ucmlud}). Besides, the 8 target datasets include CropDisease, EuroSAT, ISIC, ChestX, CASIA NIR Database (faceNIR)~\cite{CBSR}, Fish Recognition data (Fish)~\cite{Fish}, Colorectal Histology MNIST (CHM)~\cite{Medical}, Audio-Visual Vehicle (AVV) Dataset~\cite{vehicle}.

\textbf{Network structure and training settings.}
To evaluate the performance of ISSNet, we consider ResNet10~\cite{resnet10} as the student backbone. The input size is $224 \times 224$. In the Inter-Source style transfer learning, we follow the structures of~\cite{adain}. And in the cross-domain distillated consistency learning, the student model \textit{M}, followed by a classifier, is trained for 1000 epochs with a batch size of 128. And it is optimized with stochastic gradient descent (SGD). The learning rate, momentum and weight decay are set as $10^{-3}$, $0.9$ and $5 \times 10^{-4}$, respectively. In the target prototypical classifier learning phase, \textit{M} and a new classifier are optimized with SGD for 600 epochs firstly. The learning rate, momentum and weight decay are set to $10^{-2}$, $0.9$ and $10^{-3}$. Additionally, two strategies that have been shown to be effective are applied to the CDFSC task: data augmentation (DA), and label propagation (LP)~\cite{fsl:best}. DA is used in the latter two stages, and the LP is utilized in the third phase. Note that the pre-trained ResNet10 on ImageNet is not used in this work. And we only evaluate the 5-way 1-shot task on AVV cause of the extremly limit data.

\vspace{-0.1cm} 
\subsection{Effectiveness of Inter-Source style transfer learning}
\begin{figure}
  \centering
  \includegraphics[width=0.8\linewidth]{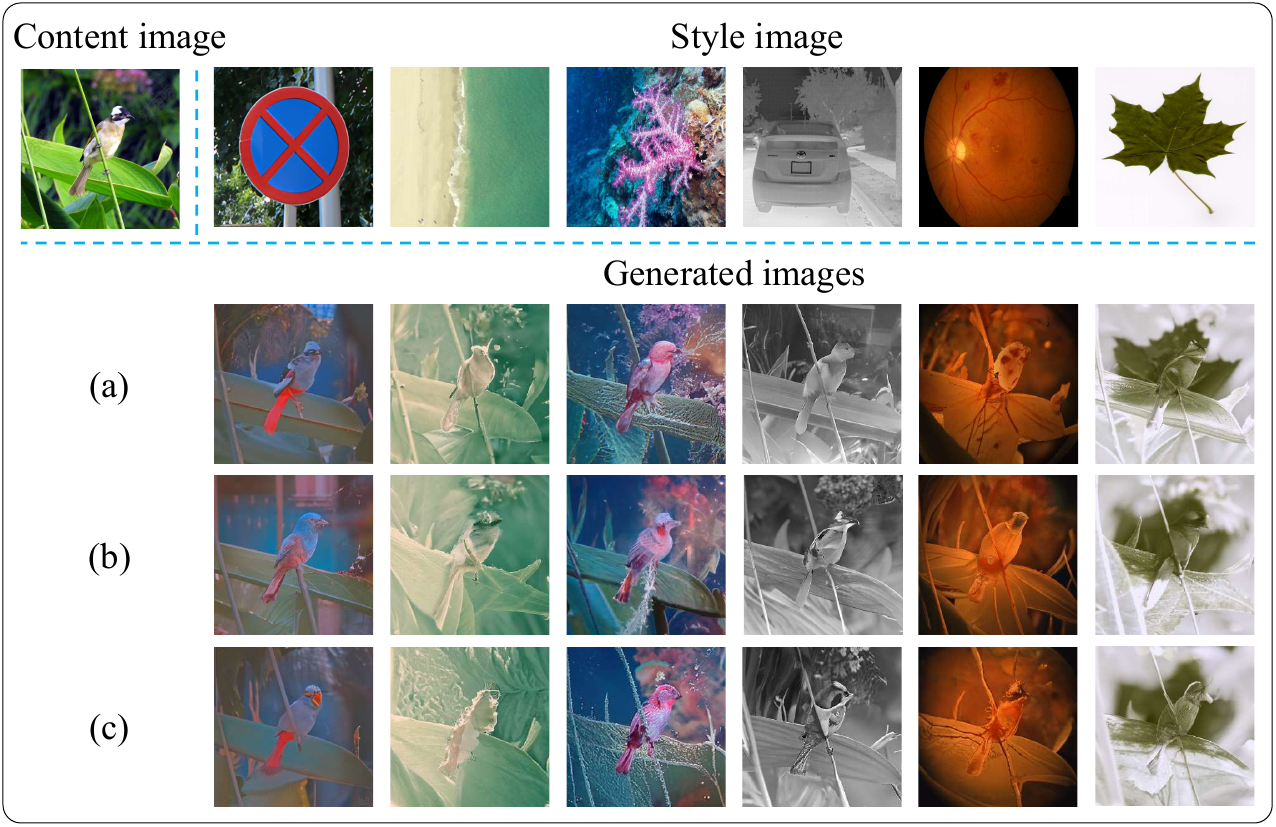}
  \vspace{-0.3cm}
    \caption{Visualization of images generated by style transfer network. The generated images consists of the contents of labelled source and the styles of unlabelled source. (a), (b), and (c) show three generated images for each content and style image pair.}
    \vspace{-0.5cm}
    \label{fig:adain}
\end{figure}

We evaluate the effects of Inter-Source style transfer learning, and visualize the images generated by \textit{A} in Figure~\ref{fig:adain}. Sample of $\mathcal{D}_{sl}$ are regard as the content images, and that of $\mathcal{D}_{su}$ are the style images. The generated images $\mathcal{D}_{al}$ combine the contents from $\mathcal{D}_{sl}$ and the styles from $\mathcal{D}_{su}$.  In Figure~\ref{fig:adain}, we display the three $\mathcal{D}_{al}$ for each pair of ($\mathcal{D}_{sl}$, $\mathcal{D}_{su}$). We can know that $\mathcal{D}_{al}$ always keep same silhouettes with $\mathcal{D}_{sl}$, while the styles similar to $\mathcal{D}_{su}$. Therefore, with the help of $\mathcal{D}_{su}$, $\mathcal{D}_{al}$ can expand the distribution of $\mathcal{D}_{sl}$, helping to improve the model generalization ability.

 \begin{table*}
 \vspace{-0.5cm}
 \caption{Results of ISSNet on 8 target domains about 5-way 1-shot and 5-way 5-shot tasks. The results show that our method obtains a competitive performance. The bold \textbf{black}, \textcolor{blue}{\textbf{blue}} indicate the first and the second best performance on each target domain, respectively.}
 \vspace{-0.5cm}
  \begin{center}
    {\tiny{
    \setlength{\tabcolsep}{1.0mm}{
\begin{tabular}{lcccccccc|ccccccc}
\toprule
\multirow{2}*{Methods}  & \multicolumn{8}{c|}{5-way 1-shot} & \multicolumn{7}{c}{5-way 5-shot} \\
\cline{2-16}
~  & CropDisease & EuroSAT & ISIC & ChestX & faceNIR & Fish & CHM & AVV & CropDisease & EuroSAT & ISIC & ChestX & faceNIR & Fish & CHM \\
\midrule
Finetune~\cite{resnet10} & 61.56$\pm$0.90 & 49.34$\pm$0.85 & 30.80$\pm$0.59 & 21.88$\pm$0.38 & 87.26$\pm$0.75 & 56.92$\pm$0.91 & 60.18$\pm$0.78 & 32.66$\pm$0.86 & 89.25$\pm$0.51 & 79.08$\pm$0.61 & 48.11$\pm$0.64 & 25.97$\pm$0.41 & 98.75$\pm$0.45 & 86.18$\pm$0.67 & 76.97$\pm$0.69  \\

ProtoNet~\cite{protonet} & 51.22$\pm$0.50 & 52.93$\pm$0.50 & 29.20$\pm$0.30 & 21.57$\pm$0.20  & 41.69$\pm$0.45 & 47.05$\pm$0.49 & 51.82$\pm$0.49 & 38.36$\pm$0.46 & 79.72$\pm$0.67 & 73.29$\pm$0.71 & 39.57$\pm$0.57 & 24.05$\pm$1.01  & 73.34$\pm$0.40 & 72.77$\pm$0.44 & 68.66$\pm$0.41  \\

MatchingNet~\cite{matchingnet} & 48.47$\pm$1.01 &  50.67$\pm$0.88 & 29.46$\pm$0.56 & 20.91$\pm$0.30   & 48.49$\pm$0.49 & 51.18$\pm$0.50 & 54.97$\pm$0.49 & 39.05$\pm$0.46 & 58.73$\pm$1.03 & 66.80$\pm$0.76 &	34.38$\pm$0.52 & 22.56$\pm$0.36  & 75.33$\pm$0.40 & 73.46$\pm$0.45 & 68.95$\pm$0.43 \\

RelationNet~\cite{relationnet} & 56.18$\pm$0.85 & 56.28$\pm$0.82 & 29.69$\pm$0.60 & 21.94$\pm$0.42  & 44.96$\pm$0.47 & 49.16$\pm$0.51 & 53.63$\pm$0.50 & 38.65$\pm$0.45 & 75.33$\pm$0.71 & 68.08$\pm$0.69 & 37.07$\pm$0.52 & 24.34$\pm$0.41  & 70.71$\pm$0.42 & 72.58$\pm$0.44 & 69.47$\pm$0.44 \\

GNN~\cite{gnn} & 59.19$\pm$0.63 & 54.61$\pm$0.50 & 30.14$\pm$0.30 & 21.94$\pm$0.24   & 48.20$\pm$0.58 & 51.20$\pm$0.59 & 51.30$\pm$0.56 & 37.29$\pm$0.51 & 83.12$\pm$0.40 & 78.69$\pm$0.42 & 42.54$\pm$0.38 & 23.87$\pm$0.19  & 80.40$\pm$0.47 & 79.79$\pm$0.51 & 73.01$\pm$0.52 \\

FWT~\cite{cdfsl} & 66.36$\pm$1.04 & 62.36$\pm$1.05 & 31.58$\pm$0.67 & 22.04$\pm$0.44  & 40.95$\pm$0.45 & 47.92$\pm$0.51 & 52.35$\pm$0.48 & 37.85$\pm$0.45 & 87.11$\pm$0.67 & 83.01$\pm$0.79 & 43.17$\pm$0.70 & 25.18$\pm$0.45   & 73.15$\pm$0.40 & 73.88$\pm$0.42 & 69.20$\pm$0.42 \\

LRP~\cite{lrp} & 59.23$\pm$0.50 & 54.99$\pm$0.50 & 30.94$\pm$0.30 & 22.11$\pm$0.20  & 45.63$\pm$0.48 & 49.58$\pm$0.52 & 53.56$\pm$0.51 & 38.47$\pm$0.47 & 86.15$\pm$0.40 & 77.14$\pm$0.40 & 44.14$\pm$0.40 & 24.53$\pm$0.30   & 72.69$\pm$0.41 & 73.87$\pm$0.44 & 69.22$\pm$0.44  \\

STARTUP~\cite{fsl:STARTUP}  & 70.09$\pm$0.86 & 64.32$\pm$0.87 & 29.73$\pm$0.51 & 22.10$\pm$0.40 & 90.22$\pm$0.70 & 65.15$\pm$0.98 & 62.03$\pm$0.85 & 37.14$\pm$0.46 & 90.81$\pm$0.49 & 82.29$\pm$0.10 & 43.55$\pm$0.56 & 26.03$\pm$0.44 & 99.27$\pm$0.16 & 87.13$\pm$0.67 & 81.94$\pm$0.58  \\

PATA~\cite{fsl:adversarial} & 67.47$\pm$0.50 & 61.35$\pm$0.50 & 33.21$\pm$0.40 & 22.10$\pm$0.20   & 37.74$\pm$0.43 & 49.14$\pm$0.51 & 49.59$\pm$0.46 & 37.48$\pm$0.46 & 90.59$\pm$0.30 & \textcolor{blue}{\textbf{83.75$\pm$0.40}} & 44.91$\pm$0.40 & 24.32$\pm$0.40   & 71.38$\pm$0.40 & 73.13$\pm$0.43 & 67.69$\pm$0.43 \\

RATA~\cite{fsl:adversarial}  & \textbf{74.61$\pm$0.50} & \textbf{66.18$\pm$0.50} & 32.96$\pm$0.30 & 22.24$\pm$0.20  & 43.78$\pm$0.47 & 50.64$\pm$0.52 & 53.20$\pm$0.50 & 38.09$\pm$0.46 & 90.80$\pm$0.30 & 81.92$\pm$0.30 & 46.99$\pm$0.30 & 25.69$\pm$0.20  & 71.60$\pm$0.42 & 72.08$\pm$0.45 & 67.60$\pm$0.42 \\



VDB~\cite{vdb} & 71.98$\pm$0.82 & 63.60$\pm$0.87 & 35.32$\pm$0.65 & 22.99$\pm$0.44  & - & - & - & - & 90.77$\pm$0.49 & 82.06$\pm$0.63 & 48.72$\pm$0.65 & 26.62$\pm$0.45  & - & - & - \\

MFDM~\cite{fsl:mixup} & 66.23$\pm$1.03 & 62.97$\pm$1.01 & 32.48$\pm$0.64 & 22.26$\pm$0.45 & 71.00$\pm$1.51 & 65.70$\pm$2.81 & 63.00$\pm$2.64 & \textbf{65.78$\pm$3.25} & 87.27$\pm$0.69 & 80.48$\pm$0.79 & 44.28$\pm$0.66 & 24.52$\pm$0.44  & 91.67$\pm$1.55 & 90.53$\pm$1.53 & \textbf{90.92$\pm$1.56} \\

\midrule
ISSNet (CE)        & 73.53$\pm$0.16 & 63.46$\pm$0.38 & 36.04$\pm$0.55 & 22.97$\pm$0.50 & 93.39$\pm$0.63 & 68.54$\pm$0.17 & 63.89$\pm$0.73 & 41.15$\pm$1.10 & 93.00$\pm$0.32 & 82.59$\pm$0.39 & 47.88$\pm$0.84 & 28.36$\pm$0.90  & 99.10$\pm$0.10 & 85.95$\pm$0.69 & 79.97$\pm$0.64  \\
ISSNet (BSR)   & 73.60$\pm$0.60 & 64.57$\pm$0.88 & 36.06$\pm$0.49 & 23.03$\pm$0.62 & 93.41$\pm$0.52 & \textcolor{blue}{\textbf{68.55$\pm$0.85}} & 64.70$\pm$0.58 & 54.91$\pm$1.07 & 93.95$\pm$0.53 & 83.54$\pm$0.53 & 51.69$\pm$0.62 & \textcolor{blue}{\textbf{28.80$\pm$0.88}}  & 99.27$\pm$0.08 & \textbf{94.19$\pm$0.64} & 82.29$\pm$0.60  \\
ISSNet (KL+CE)    & 73.58 $\pm$0.47 & 65.16$\pm$0.86 & \textcolor{blue}{\textbf{36.07$\pm$0.48}} & \textcolor{blue}{\textbf{23.30$\pm$0.62}} & \textcolor{blue}{\textbf{93.65$\pm$0.17}} & 68.50$\pm$0.63 & \textbf{65.90$\pm$0.89} & 54.47$\pm$1.52 & \textcolor{blue}{\textbf{94.10$\pm$0.70}} & 83.66$\pm$0.89 & \textcolor{blue}{\textbf{51.83$\pm$0.93}} & 28.59$\pm$0.95 & \textcolor{blue}{\textbf{99.43$\pm$0.49}} & 94.06$\pm$0.24 & 82.50$\pm$0.50  \\
ISSNet (KL+BSR)     & \textcolor{blue}{\textbf{74.48$\pm$0.28}} & \textcolor{blue}{\textbf{65.47$\pm$0.20}} & \textbf{36.63$\pm$0.26} & \textbf{23.38$\pm$0.52} & \textbf{94.39$\pm$0.37} & \textbf{68.56$\pm$0.33} & \textcolor{blue}{\textbf{65.75$\pm$0.32}} & \textcolor{blue}{\textbf{55.37$\pm$1.26}} & \textbf{94.36$\pm$0.24} & \textbf{83.84$\pm$0.75} & \textbf{51.89$\pm$0.51} & \textbf{28.81$\pm$0.19}  & \textbf{99.74$\pm$0.12} & \textcolor{blue}{\textbf{94.14$\pm$0.49}} & \textcolor{blue}{\textbf{83.63$\pm$0.69}} \\
\bottomrule
\end{tabular}}
}}
\end{center}
\vspace{-0.5cm} 
\label{tab:compare}
\end{table*}

 \vspace{-0.1cm} 
 \subsection{CDFSC ability analysis}
To verify the CDFSC effects of ISSNet, We evaluated the FSC performance on 8 different target datasets. We compare ISSNet to the baseline~\cite{resnet10}, three classical methods MatchingNet~\cite{matchingnet}, RelationNet~\cite{relationnet}, ProtoNet~\cite{protonet}, and several SOTA methods GNN~\cite{gnn}, FWT~\cite{cdfsl}, LRP~\cite{lrp}, STARTUP~\cite{fsl:STARTUP}, RelationNet+ATA (RATA)~\cite{fsl:adversarial}, ProtoNet+ATA (PATA)~\cite{fsl:adversarial}, VDB~\cite{vdb}, and Meta-FDMixup (MFDM)~\cite{fsl:mixup}. Table~\ref{tab:compare} shows the performance of ISSNet on 8 target domains.

The table shows the proposed ISSNet obtains the top two results on 8 target domains. Besides, MFDM~\cite{fsl:mixup} and STARTUP~\cite{fsl:STARTUP} show the competitive results on several datasets because they both introduce the target datasets into the source training phase. FMDM~\cite{fsl:mixup} and STARTUP~\cite{fsl:STARTUP} introduce the labelled target datasets and unlabelled target datasets into the training phase, respectively. Due to the data in AVV is blurry, the introduction of labelled target can largely help model adaptation. Hence, FMDM~\cite{fsl:mixup} performs best on AVV for 5-way 1-shot task. Similar to 5-way 1-shot task, MFDM~\cite{fsl:mixup} and STARTUP~\cite{fsl:STARTUP} indicate the particularly stunning effect on 5-shot because introducing the target datasets into the training phase.

 \vspace{-0.1cm}
\subsection{Ablation study}
\textbf{Performance of losses.}
Recently, several works utilize KL loss or introduce the batch spectral regularization (BSR)  into CE loss to improve the CDFSC performances. Therefore, we compare the performance of ISSNet optimazed with these losses on 5-way 1-shot and 5-way 5-shot task, as shown in the bottom of Table~\ref{tab:compare}. 

The table shows "KL+CE" and "KL+BSR" performs best on the most of target datasets. We analyze that this is because the KL loss constrains the similarity of the outputs of student and teacher networks, making the model further generalize to the different styles. Furthermore, the results also show that the BSR loss performs better than CE loss, which means the BSR loss is more suitable for solving CDFSC problem than CE loss. In general, "KL+BSR" obtains the best performance.

\begin{figure}
  \centering
  \includegraphics[width=0.9\linewidth]{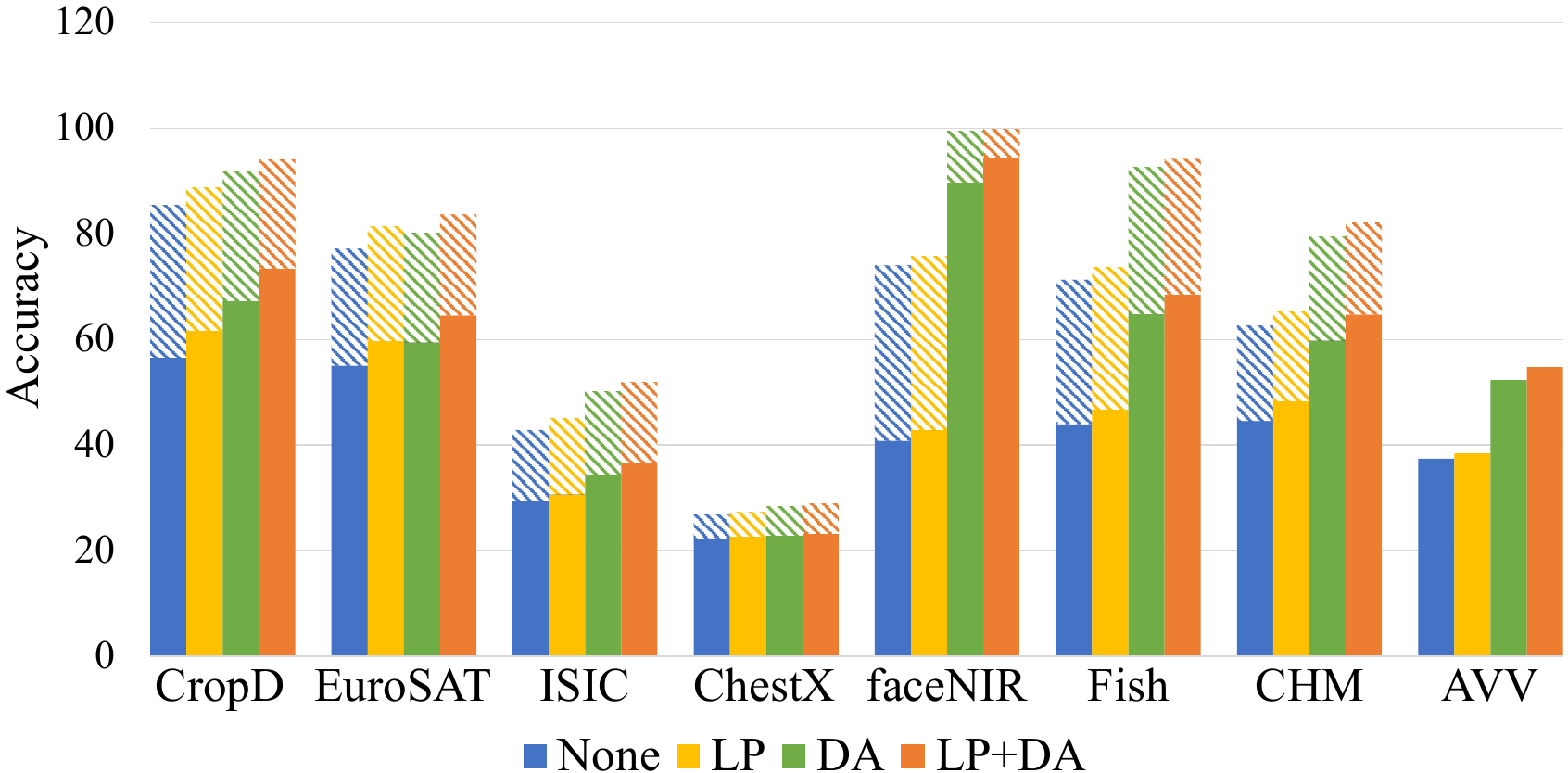}
  \vspace{-0.3cm}
    \caption{The ISSNet performances on 8 different target datasets for 5-way 1-shot (solid fill) and 5-way 5-shot (stripe fill) tasks.``CropD" indicates the dataset ``CropDiseases".}
    \vspace{-0.5cm}
    \label{fig:strategy}
\end{figure}

\textbf{Performance of strategies.}
We introduce two strategies, data augmentation (DA) and label propagation (LP), to improve the CDFSC performance. Figure~\ref{fig:strategy} shows the results of the introduction of strategies, in which `None' means do not use any strategies. We can know from figure that the introduce of DA can plenty improve the CDFSC performance on all of the target domains, while only introduce the LP have not greatly improve the  performance. Moreover, the introduction of both LP and DA (LP+DA) achieves the best effects.

\section{Conclusions}
\label{sec:5}
This paper introduces multiple source domains into the model training with only one source is fully labelled while the rest sources remain unlabelled.  To address the challenge in this situation, this paper first explores a benchmark. And then an Inter-Source Stylization network (ISSNet) is proposed to improve the generalization ability of model, including Inter-Source style transfer learning, cross-domain distillated consistency learning, and target prototypical classifier learning. In general, this paper introduces multiple source domains for CDFSC studies without increasing labeling costs, and proposes ISSNet to address the corresponding challenges. Evaluation on 8 datasets show that ISSNet significantly improve the CDFSC performance.

\vspace{-0.2cm}
\section*{Acknowledgements}
This work was partially supported by National Key R\&D Program of China No. 2021YFB3100800, the Academy of Finland under grant 331883.

\bibliographystyle{IEEEbib}
\bibliography{Template}

\begin{thebibliography}{10}

\bibitem{fsl:survey}
Y~Wang, Q~Yao, J~T Kwok, and L~M Ni,
\newblock ``Generalizing from a few examples: A survey on few-shot learning,''
\newblock {\em ACM Computing Surveys}, , no. 3, pp. 1--34, 2020.

\bibitem{fsl:classification}
W~Chen, Y~Liu, Z~Kira, Y~F Wang, and J~Huang,
\newblock ``A closer look at few-shot classification,''
\newblock in {\em ICLR}, 2019.

\bibitem{cdfsl}
H~Tseng, H~Lee, J~Huang, and M~Yang,
\newblock ``Cross-domain few-shot classification via learned feature-wise
  transformation,''
\newblock in {\em ICLR}, 2020.

\bibitem{resnet10}
Y~Guo, N~C Codella, L~Karlinsky, J~V Codella, J~R Smith, K~Saenko, T~Rosing,
  and R~Feris,
\newblock ``A broader study of cross-domain few-shot learning,''
\newblock in {\em ECCV}, 2020, pp. 124--141.

\bibitem{fsl:adversarial}
H~Wang and Z~Deng,
\newblock ``Cross-domain few-shot classification via adversarial task
  augmentation,''
\newblock {\em IJCAI}, 2021.

\bibitem{fsl:ensemble}
T~Adler, J~Brandstetter, M~Widrich, A~Mayr, D~Kreil, M~Kopp, G~Klambauer, and
  S~Hochreiter,
\newblock ``Cross-domain few-shot learning by representation fusion,''
\newblock {\em arXiv preprint arXiv:2010.06498}, 2020.

\bibitem{fsl:mixup}
Y~Fu, Y~Fu, and Y~Jiang,
\newblock ``Meta-fdmixup: Cross-domain few-shot learning guided by labeled
  target data,''
\newblock in {\em ACM MM}, 2021, pp. 5326--5334.

\bibitem{cdfsl:Tian}
P~Tian and Y~Gao,
\newblock ``Improving the generalization of meta-learning on unseen domains via
  adversarial shift,''
\newblock {\em arXiv preprint arXiv:2107.11056}, 2021.

\bibitem{mcd1}
W~Li, X~Liu, and H~Bilen,
\newblock ``Universal representation learning from multiple domains for
  few-shot classification,''
\newblock in {\em IEEE/CVF ICCV}, 2021, pp. 9526--9535.

\bibitem{mcd2}
W~Li, X~Liu, and H~Bilen,
\newblock ``Cross-domain few-shot learning with task-specific adapters,''
\newblock in {\em IEEE/CVF CVPR}, 2022, pp. 7161--7170.

\bibitem{cnsn}
Z~Tang, Y~Gao, Y~Zhu, Z~Zhang, M~Li, and D~N Metaxas,
\newblock ``Crossnorm and selfnorm for generalization under distribution
  shifts,''
\newblock in {\em IEEE/CVF ICCV}, 2021, pp. 52--61.

\bibitem{adain}
X~Huang and S~Belongie,
\newblock ``Arbitrary style transfer in real-time with adaptive instance
  normalization,''
\newblock in {\em IEEE/CVF ICCV}, 2017, pp. 1501--1510.

\bibitem{imagenet}
J~Deng, W~Dong, R~Socher, L~Li, K~Li, and F~Li,
\newblock ``Imagenet: A large-scale hierarchical image database,''
\newblock in {\em IEEE/CVF CVPR}, 2009, pp. 248--255.

\bibitem{matchingnet}
O~Vinyals, C~Blundell, T~Lillicrap, D~Wierstra, et~al.,
\newblock ``Matching networks for one shot learning,''
\newblock {\em NIPS}, 2016.

\bibitem{eyeball}
T~Kauppi, V~Kalesnykiene, JK~Kamarainen, L~Lensu, I~Sorri, A~Raninen,
  R~Voutilainen, H~Uusitalo, H~K{\"a}lvi{\"a}inen, and J~Pietil{\"a},
\newblock ``The diaretdb1 diabetic retinopathy database and evaluation
  protocol,''
\newblock in {\em BMVC}, 2007, pp. 1--10.

\bibitem{leaf}
Oskar S{\"o}derkvist,
\newblock ``Computer vision classification of leaves from swedish trees,''
  2001.

\bibitem{SUIM}
M~J Islam, C~Edge, Y~Xiao, P~Luo, M~Mehtaz, C~Morse, S~S Enan, and J~Sattar,
\newblock ``Semantic segmentation of underwater imagery: Dataset and
  benchmark,''
\newblock in {\em IEEE/RSJ ICIRS}, 2020, pp. 1769--1776.

\bibitem{traffic}
J~Zhang, W~Wang, C~Lu, J~Wang, and A~K Sangaiah,
\newblock ``Lightweight deep network for traffic sign classification,''
\newblock {\em Annals of Telecommunications}, pp. 369--379, 2020.

\bibitem{fsl:ucmlud}
Y~Yang and S~Newsam,
\newblock ``Bag-of-visual-words and spatial extensions for land-use
  classification,''
\newblock in {\em International Conference on Advances in Geographic
  Information Systems}, 2010, pp. 270--279.

\bibitem{CBSR}
S~Z Li, R~Chu, S~Liao, and L~Zhang,
\newblock ``Illumination invariant face recognition using near-infrared
  images,''
\newblock {\em IEEE TPAMI}, , no. 4, pp. 627--639, 2007.

\bibitem{Fish}
O~Ulucan, D~Karakaya, and M~Turkan,
\newblock ``A large-scale dataset for fish segmentation and classification,''
\newblock in {\em IISAC}, 2020, pp. 1--5.

\bibitem{Medical}
M~L Fravolini, M~Belal, B~Palumbo, and J~N Kather,
\newblock ``Dimensionality reduction strategies for cnn-based classification of
  histopathological images,''
\newblock {\em Intelligent Interactive Multimedia Systems and Services}, p.~21,
  2017.

\bibitem{vehicle}
T~Wang and Z~Zhu,
\newblock ``Real time moving vehicle detection and reconstruction for improving
  classification,''
\newblock in {\em IEEE Workshop on the Applications of Computer Vision (WACV)},
  2012, pp. 497--502.

\bibitem{fsl:best}
B~Liu, Z~Zhao, Z~Li, J~Jiang, Y~Guo, and J~Ye,
\newblock ``Feature transformation ensemble model with batch spectral
  regularization for cross-domain few-shot classification,''
\newblock {\em arXiv preprint arXiv:2005.08463}, 2020.

\bibitem{protonet}
J~Snell, K~Swersky, and R~Zemel,
\newblock ``Prototypical networks for few-shot learning,''
\newblock in {\em NIPS}, 2017, pp. 4080--4090.

\bibitem{relationnet}
F~Sung, Y~Yang, L~Zhang, T~Xiang, P~HS Torr, and T~M Hospedales,
\newblock ``Learning to compare: Relation network for few-shot learning,''
\newblock in {\em IEEE/CVF CVPR}, 2018, pp. 1199--1208.

\bibitem{gnn}
V~Garcia and J~Bruna,
\newblock ``Few-shot learning with graph neural networks,''
\newblock in {\em ICLR}, 2018.

\bibitem{lrp}
J~Sun, S~Lapuschkin, W~Samek, Y~Zhao, N~Cheung, and A~Binder,
\newblock ``Explanation-guided training for cross-domain few-shot
  classification,''
\newblock in {\em ICPR}, 2021, pp. 7609--7616.

\bibitem{fsl:STARTUP}
C~P Phoo and B~Hariharan,
\newblock ``Self-training for few-shot transfer across extreme task
  differences,''
\newblock in {\em ICLR}, 2020.

\bibitem{vdb}
M~Yazdanpanah and P~Moradi,
\newblock ``Visual domain bridge: A source-free domain adaptation for
  cross-domain few-shot learning,''
\newblock in {\em Proceedings of the IEEE/CVF CVPR}, 2022, pp. 2868--2877.

\end{thebibliography}

\end{document}